\documentclass{article}

\usepackage[nonatbib,preprint]{neurips_2023}

\usepackage[utf8]{inputenc} 
\usepackage[T1]{fontenc}    
\usepackage{hyperref}       
\usepackage{url}            
\usepackage{booktabs}       
\usepackage{amsfonts}       
\usepackage{nicefrac}       
\usepackage{microtype}      
\usepackage{xcolor}         
\usepackage{amsmath}
\usepackage{graphicx}

\title{Nowcasting day-ahead marginal emissions using multi-headed CNNs and deep generative models}

\author{%
  Dhruv Suri \\
  Energy Science and Engineering\\
  Stanford University\\
  Stanford, CA 94305 \\
  \texttt{surid@stanford.edu} \\
    \And
  Anela Arifi \\
  EIPER \\
  Stanford University \\
  Stanford, CA 94305 \\
  \texttt{anelaa@stanford.edu} \\
  \AND
  In\^es Azevedo \\
  Energy Science and Engineering\\
  Stanford University\\
  Stanford, CA 94305 \\
  \texttt{iazevedo@stanford.edu}
}

\begin{document}

\maketitle

\begin{abstract}
Nowcasting day-ahead marginal emissions factors is increasingly important for power systems with high flexibility and penetration of distributed energy resources. With a significant share of firm generation from natural gas and coal power plants, forecasting day-ahead emissions in the current energy system has been widely studied. In contrast, as we shift to an energy system characterized by flexible power markets, dispatchable sources, and competing low-cost generation such as large-scale battery or hydrogen storage, system operators will be able to choose from a mix of different generation as well as emission pathways. To fully develop the emissions implications of a given dispatch schedule, we need a near real-time workflow with two layers. The first layer is a market model that continuously solves a security-constrained economic dispatch model. The second layer determines the marginal emissions based on the output of the market model, which is the subject of this paper. We propose using multi-headed convolutional neural networks to generate day-ahead forecasts of marginal and average emissions for a given independent system operator.
\end{abstract}

\section{Introduction}

In the U.S., electric-power systems are characterized by generation mixes dominated with firm generation resources such as natural gas, coal, hydropower and nuclear. Forecasting scenarios of day-ahead emissions with largely firm resources is relatively simple although not trivial. This complexity is due to competitive power markets, heterogeneity in nodal vs wholesale power market design, and other factors such as the relative size of the spot market compared to the day-ahead market. In contrast, as we shift to an energy system characterized by flexible power markets, dispatchable sources, and competing low-cost generation such as large-scale battery or hydrogen storage, system operators will be able to choose from a mix of different generation as well as emission pathways. 

With electrification, multiple sectors of the economy rely on knowing real-time market prices as well as emissions to make decisions. For example, enterprise fleet operators that account for Scope 2 emissions charge their fleet of electric vehicles when the emissions intensity of the grid is the lowest. Thus, having real-time or near real-time access to these key parameters is becoming increasingly important. To fully develop the emissions implications of a given dispatch schedule, we need a near real-time workflow with two layers. The first layer is a market model that continuously solves a security-constrained economic dispatch model. The second layer determines the marginal emissions based on the output of the market model, which is the subject of this paper.

Marginal emissions and marginal emissions factors help quantify the health, environmental, and climate change impacts caused by changes in marginal net electricity consumption, which could result from new technologies or policies \cite{deetjen2019reduced, siler2012marginal, thind2017marginal, sengupta2022current, yuksel2016effect}. In our analysis, we focus on predicting day-ahead marginal emissions given a system-level forecast of demand. This also allows us to conduct a sensitivity analysis on the computed emissions by marginally increasing the day-ahead forecasts to characterize the exact impact on the resulting generation mix. Most studies and capacity expansion analyses focus on the average emissions intensity of a given generation mix \cite{bilgili2016dynamic, saidi2020impact, bhattacharya2017dynamic}. While, the average emissions intensity provides a systems-level metric, it does not yield any information in the event that the forecasts are inaccurate. The marginal emissions intensity, however, characterizes where the extra unit of generation is coming from so that we can choose from a variety of dispatch alternatives.

We propose predicting day-ahead emissions, marginal emissions, and marginal emissions intensity using day-ahead forecast of hourly demand or hourly generation by fuel type. Convolutional neural networks (CNNs) are a type of neural network that was initially designed to handle image data \cite{lecun1995convolutional}. The ability of CNNs to automatically extract features from raw input data can be applied very effectively to time series forecasting problems \cite{dorffner1996neural}. CNNs support multivariate inputs, multi-step outputs, and when coupled with Long Short Term Memory (LSTM) networks, support the efficient learning of temporal dependencies \cite{brownlee2018deep,gamboa2017deep}.

\section{Methodology}

\subsection{Data}

We obtain hourly demand, demand forecasts, generation by fuel type, and information on net imports at the balancing authority level from the U.S. Energy Information Administration (EIA). This dataset contains real-time hourly generation by fuel type (coal, natural gas, solar, wind, etc.) along with the corresponding emissions, demand forecast, and observed system demand. For the preliminary version of results presented in this paper, we obtain an hourly dataset for the California Independent System Operator (CAISO) for the period 01 January 2023 through 31 August 2023.

\subsection{Methodology}

For every 24-hour period, the average emissions factor for $\text{CO}_2$ is taken as the ratio of the total emissions to the total electricity generated. In contrast, the marginal emissions, $\Delta E_{t}^{\text{CO}_2}$, and marginal generation, $\Delta G_{t}^{\text{fossil}}$, are computed using equations \ref{marginal_emissions} and \ref{marginal_generation}. The marginal emissions intensity is the ratio of the marginal emissions to the marginal generation for each timestamp $t$.

\begin{equation}
    \Delta E_{t}^{\text{CO}_2} = \Delta E_{t}^{\text{CO}_2} - \Delta E_{t-1}^{\text{CO}_2}
    \label{marginal_emissions}
\end{equation}

\begin{equation}
    \Delta G_{t}^{\text{fossil}} = \Delta G_{t}^{\text{fossil}} - \Delta G_{t-1}^{\text{fossil}}
    \label{marginal_generation}
\end{equation}

For the initial phase of the study, we develop a multi-input, multi-step CNN architecture. The key parameters to tune and optimize are the number of kernels, strides, pooling layer and the flattening. Later, we plan to configure this architecture for use with an LSTM network to parse the temporal nature of the inputs and also couple an autoencoder for dimensionality reduction. Using the system-level day-ahead forecast, previous day demand, observed emissions, marginal emissions, and marginal demand, we train the CNN model with observed emissions as the endogenous variable. Other potential exogenous variables to consider are day of week/month/year, hour of week, an indicator variable for national holidays, as well as meteorological information. Using meteorological data to generate the day-ahead forecast will require forecasts of day-ahead precipitation, temperature, wind speed and irradiance, which have their associated uncertainty and error bounds \cite{biswas2017optimal}. 

We plan to integrate Google Deepmind's GraphCast weather model to make these day-ahead predictions. GraphCast is an open-source weather model developed by Google and has shown to have considerably high accuracy compared to the European Centre for Medium-Range Weather Forecasts (ECMWF)'s new High REsolution forecast (HRES) in various regions of the world \cite{lam2022graphcast}.

\section{Proposed Direction}

\begin{figure}[h!]
  \centering
  \includegraphics[width=1\linewidth]{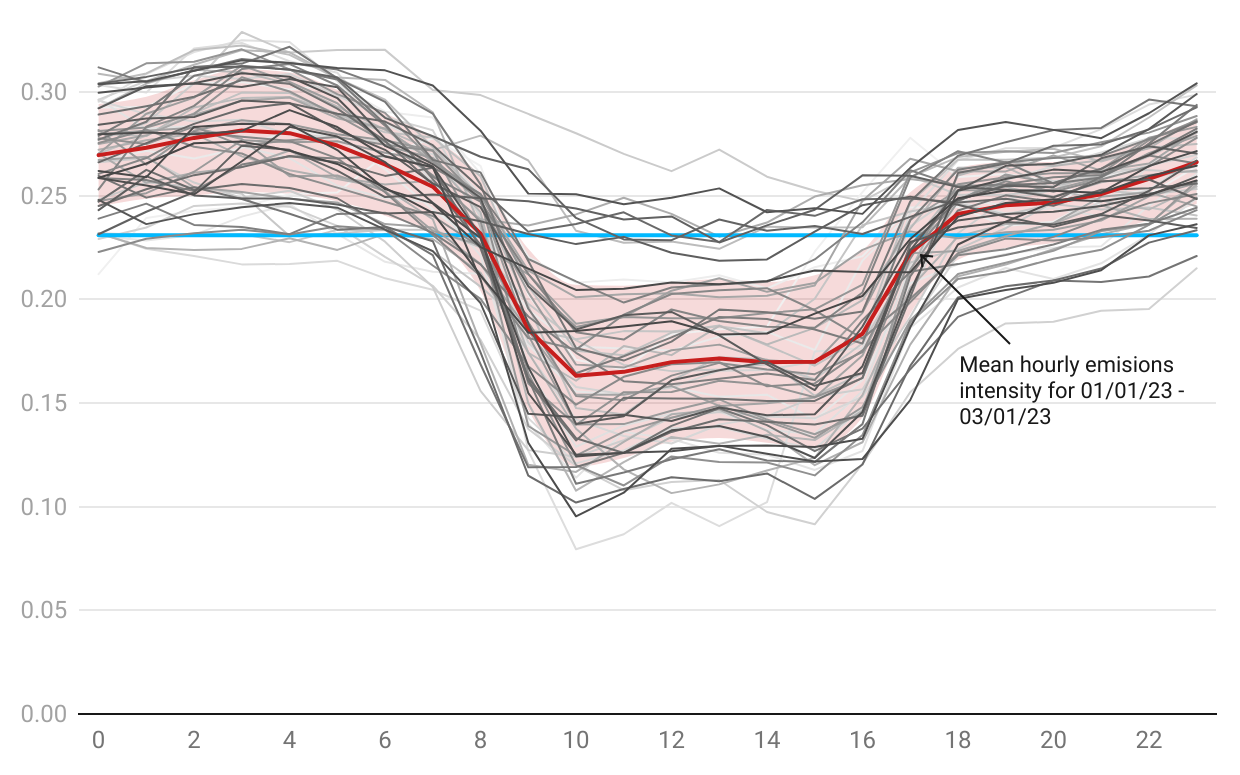}
  \caption{The characteristic 'duck curve' for CAISO, January and February 2023. Hourly grid intensity for CAISO for the time period 01/01/2023 through 03/01/2023. The red line indicates the mean intensity for the 60-day period whereas the blue line shows the mean or average emissions intensity for the same period. The $y$ axis shows the emissions intensity in tonnes of $\text{CO}_2$ produced per MWh of generated electricity and the $x$ represents the hour of day.}
  \label{JF_sample}
\end{figure}

Figure \ref{JF_sample} demonstrates that the emissions intensity varies as a function of the hour-of-day, and is not constant for the 60-day period shown in the graph. Using a mean hourly emissions intensity depicted by the red line results in the loss of granular information that may lead to incorrectly computing the marginal emissions of a dispatch schedule. Similarly, considering an average emissions intensity value depicted by the blue line shows that we at times underestimate or overestimate emissions over the course of the day. 

For system operators to select the most appropriate dispatch schedule, more granular information is needed, and we believe that the proposed approach leveraging CNNs, LSTM and other combinatorial networks will yield significant advantages compared to naive or statistical approaches currently used.

Using methods that allow us to do day-ahead nowcasting as data continuously becomes available will further generate real-time insights for marginal emissions that enable downstream demand-side optimization strategies. Daily information does not warrant the continuous re-training of the CNN model every 24 hours although the most recent time period are continuously appended to the prediction dataset for exogenous variables. 

As we continue to advance our understanding of forecasting real-time emissions, this methodology has the potential to support a wide range of applications, from enhancing planning for ISOs to optimizing fleet charging for electric vehicles.

\bibliographystyle{unsrt}
\bibliography{neurips_2023}

\end{document}